\pdfoutput=1

\documentclass[11pt]{article}

\usepackage[final]{acl}
\usepackage[final]{acl}

\usepackage{times}
\usepackage{latexsym}
\usepackage{booktabs}
\usepackage[T1]{fontenc}
\usepackage[utf8]{inputenc}

\usepackage{microtype}

\usepackage{inconsolata}

\usepackage{graphicx}

\usepackage{amsmath}    
\usepackage{algorithm}  
\usepackage{algpseudocode}
\usepackage{multirow}
\usepackage{float}

\title{Reason from Future: Reverse Thought Chain Enhances LLM Reasoning}

\author{
Yinlong Xu$^{1}$, Yanzhao Zheng$^{2}$, Shuoshuo Sun$^{2}$, \\ Shuaihan Huang$^{2}$,
Baohua Dong$^{2}$, Hangcheng Zhu$^{2}$, Ruohui Huang$^{2}$, Gang Yu$^{2}$, Hongxia Xu$^{3,4}$\thanks{\, Corresponding authors.}, 
Jian Wu$^{1,5}$\footnotemark[1]\\
$^{1}$ College of Computer Science and Technology, Zhejiang University, Hangzhou, China\\
$^{2}$ Alibaba Group, Hangzhou, China\\
$^{3}$ State Key Laboratory of Transvascular Implantation Devices and TIDRI, Hangzhou, 310009, China\\
$^{4}$ Liangzhu Laboratory and WeDoctor Cloud, Hangzhou, 310000, China\\
$^{5}$ Zhejiang Key Laboratory of Medical Imaging Artificial Intelligence, Hangzhou, 310058, China\\
\texttt{\{xuyinlong, Einstein, Wujian2000\}@zju.edu.cn} \ \ \ \texttt{huangshuaihan@outlook.com}\\
\texttt{\{zhengyanzhao.zyz, sunshuoshuo.sss, baohua.dbh, linran.lr09, wentong, ruohai\}@taobao.com}
}

\begin{document}
\maketitle
\begin{abstract}
It has been demonstrated that carefully designed reasoning paradigms, like Chain-of-Thought~(CoT) and Tree-of-Thought~(ToT), can enhance the reasoning capabilities of small language models by detailed thinking and extensive thought searching, unbounded branching factors in the searching space create prohibitive reasoning consumption. However these methods fall into the trap of local optimum reasoning, which means the model lacks a global perspective while solving problems. We propose a novel reasoning paradigm called \textit{Reason from Future}~(RFF), which generates reasoning paths by bidirectional reasoning that combines top-down planning with bottom-up reasoning accumulation. The essence of RFF lies in its reverse reasoning mechanism, which prioritizes core logical relationships and imposes goal-oriented constraints on intermediate steps, thereby reducing the searching space and mitigating error accumulation inherent in sequential forward reasoning. Empirical evaluations across diverse experiments demonstrate that RFF outperforms conventional paradigms with higher accuracy and less searching space to solve complex tasks.
\end{abstract}

\section{Introduction}
The rapid evolution of large language models~(LLMs), fueled by breakthroughs in deep learning architectures and unprecedented datasets, has demonstrated remarkable potential across natural language processing~(NLP) and interdisciplinary applications~\cite{lee2018pre, radford2018improving, team2023gemini, sel2023algorithm}. LLMs like ChatGPT~\cite{achiam2023gpt} and Llama~\cite{dubey2024llama} exhibit human-like text generation, multilingual task execution, and emerging logical reasoning. Current scholarly investigations identify their reasoning capacity for problem decomposition as the critical determinant of functional boundaries, enabling industrial automation and academic research applications. 

\begin{figure}[t]
    \centering
    \hspace*{\fill}
    \includegraphics[width=0.5\textwidth]{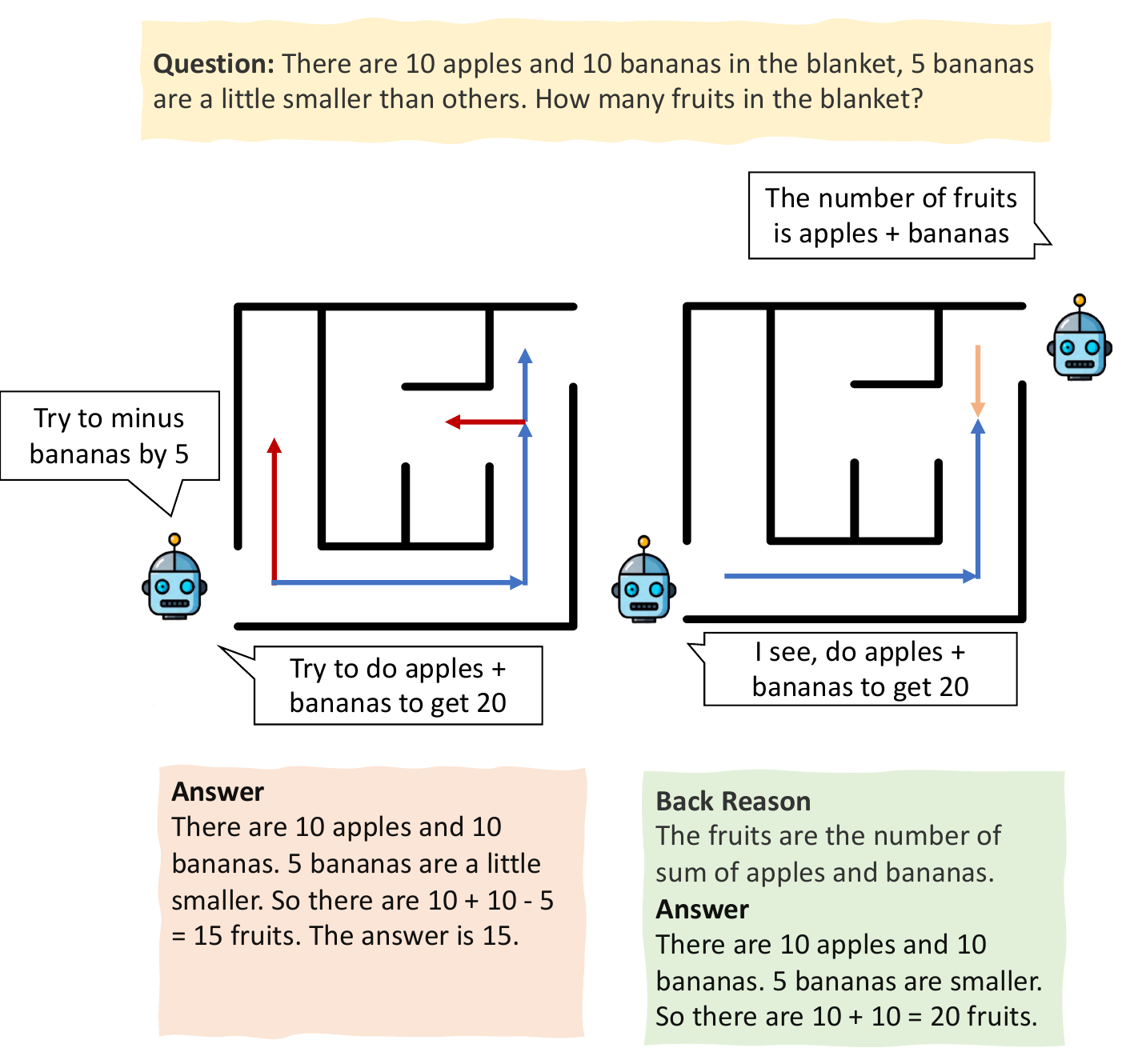}
    \hspace*{\fill}
    \caption{Comparison between simple forward reasoning~(left) and forward reasoning guided by back reasoning~(right).}
    \label{fig:inspire}
\end{figure}
 
Recent studies demonstrate that well-designed reasoning paradigms can significantly enhance LLMs' reasoning ability without additional costly and time-consuming post-training. A seminal work in this area is Chain-of-Thought~(CoT)~\citep{wei2022chain}, which pioneered the novel view that reasoning ability can be improved by designing reasoning prompts, paradigms, and examples. Tree-of-Thought~(ToT)~\cite{yao2024tree} provides a searching view to enhance the ability of complex reasoning. Progressive-Hint Prompting~(PHP)~\cite{zheng2023PHP} and Cumulative Reasoning~(CR)~\cite{zhang2023CR} asks the model to generate hints for the question before generating the answer.

Although, these reasoning paradigms, breaking down the solution into multiple steps through prompts or spatial search, can enhance the reasoning ability and coherence of the model. They tend to make the model focus on the current state, resulting in lacking explicit guidance from a global understanding of the problem and excessive exploration of redundant information, overthinking, or errors during inference~\cite{boix2023can}. 

In contrast, the way human approaches problem-solving is different. Researches have shown that humans begin by building holistic mental modeling when solving complex problems, allowing problem solvers to form a topological framework before focusing on specific details~\cite{spreng2009common,koban2021self}. This kind of cognitive prediction provides dual guidance for the subsequent solution process: forming a "cognitive road map" of the solution path at the macro-level, helping to exclude obviously unrelated branches; 
evaluation criteria are established at the micro-level so that each specific operation remains dynamically calibrated to the end goal. This global awareness allows us to avoid blindly combining superficial details and instead prioritize purposeful, contextually grounded deductions.
This suggests that modeling this local-global consistency thinking paradigm might be able to enable LLMs to strategically synthesize information, minimize irrelevant exploration, and align intermediate steps with the overarching goal.

Inspired by the maze-solving strategy of backward reasoning, where reversing the path from the endpoint accelerates discovering the solution, we propose a novel reasoning paradigm called \textbf{R}eason-\textbf{f}rom-\textbf{F}uture~(RFF) to enhance the reasoning ability of LLMs by adding reverse thinking process to guide the forward reasoning as shown in Figure~\ref{fig:inspire}. 

RFF integrates bidirectional reasoning by alternating between reverse and forward thinking to maintain solution states: the reverse reasoning generates the potential last state of the target state and sets the last state as the new target, then the forward reasoning takes a step toward the new target. The target state serves as a guide to precisely lead the forward reasoning, and the forward reasoning in turn produces more useful information to make the reverse reasoning more reasonable. We evaluate RFF in five datasets: Game of 24~\cite{yao2024tree}, GSM8K~\cite{cobbe2021training}, ASDiv~\cite{miao2021diverse}, SVAMP~\cite{patel2021nlp} MATH-500~\cite{math500}, and demonstrate significant improvements in accuracy over baseline methods. Additionally, RFF reduces the search space by constraining reasoning to target-driven states, demonstrating good efficiency. Our results highlight the potential of bidirectional, goal-aware reasoning to unlock more robust and systematic problem-solving in LLMs.

In summary, We introduce RFF, a novel self-planning reasoning paradigm to enhance the reason ability of LLMs. In which, reverse thinking and forward-thinking alternately to obtain a future perspective and narrow the solution-searching space. We conduct experiments involving four datasets to demonstrate the great performance and efficiency of RFF. And we employ two extra experiments by complicating the questions in Game of 24 and GSM8K. The results represent RFF less consuming in larger search spaces and robust thinking in variant problems.

\begin{figure*}[!htp]
    \centering
    \makebox[\textwidth][c]{\includegraphics[width=1.2\linewidth]{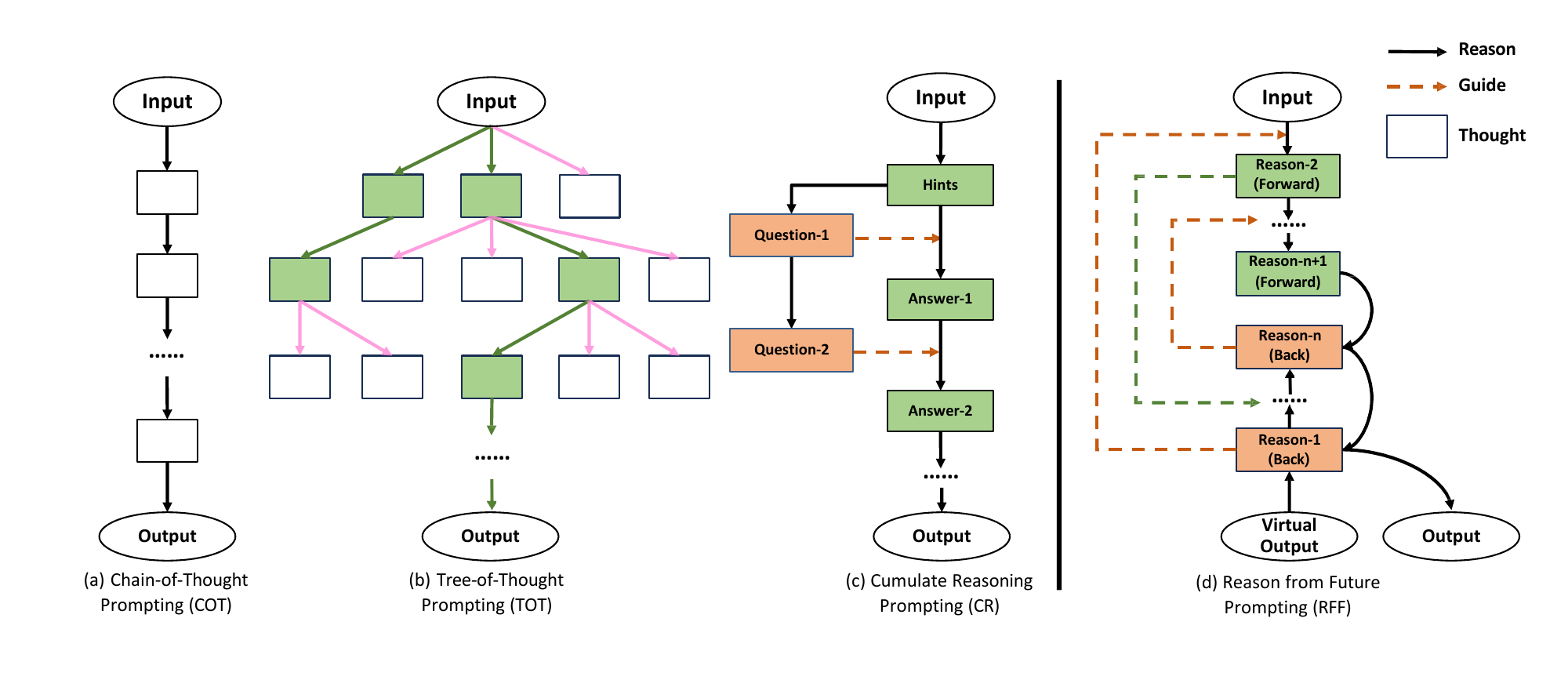}}
    \caption{Schematic illustrating various approaches to problem-solving with LLMs, and each rectangle box represents a thought. Figure\ref{fig:framework}(d) only shows the basic framework about RFF, see the concrete pipeline of two types of RFF in Algorithm ~\ref{rff-with-back}~(RFF-T) and Algorithm~\ref{rff-without-back} ~(RFF-G).}
    \label{fig:framework}
\end{figure*}

\section{Related Work}
\subsection{Chain of Thought Reasoning}
In the study of complex reasoning tasks, Chain-of-Thought~(CoT)~\cite{wei2022chain, wang2022self} prompting has emerged as a pivotal technique for significantly improving the performance of large language models~(LLMs) by explicitly generating intermediate reasoning steps. This approach enables the decomposition of problems into structured, stepwise reasoning pathways, demonstrating particular efficacy in mathematical and logical domains. Recent advancements extend CoT through symbolic formalization~(e.g., Symbolic CoT~\cite{xu2024faithful}), which incorporates formal logic systems to enhance both reliability and interpretability by grounding reasoning in rigorous symbolic frameworks. Critical analyses, however, reveal potential limitations where models may exploit computational redundancy rather than genuine reasoning in extended CoT steps, prompting discussions about mechanistic transparency.

\subsection{Search Reasoning}
In the domain of search-based reasoning for large language models, the Tree-of-Thought~(ToT)~\cite{yao2024tree} framework introduces backtracking capabilities within multi-path decision structures, enabling systematic exploration of diverse solution trajectories. This approach proves particularly effective for complex tasks requiring iterative hypothesis generation and validation. 
Monte Carlo Tree Search~(MCTS)~\cite{swiechowski2023monte} strengthens online decision-making robustness through simulating and evaluating long-term rewards of candidate paths, demonstrating strengths in reinforcement learning and dynamic programming scenarios. 
AOT~\cite{sel2023algorithm} improves TOT by introducing a step-evaluation mechanism into the reasoning process, which helps LLM to prune the less likely searching route thus reduce the searching space. AOT+~\cite{aot+}, as the upgrade of AOT, adds a fine-grained backtracking mechanism by labeling each steps, which further reduces the reasoning consumption on an error searching route. However, both AOT and AOT+ get their global perspectives by continue exploring reasoning and degrade to normal TOT when exploring their first reasoning route, which may lead to a random search in each step and may miss the correct searching route.

Recent innovations also bridge reasoning with executable action, exemplified by frameworks like LATS~(Language Agent Tree Search)~\cite{zhou2023language}. 
By unifying hierarchical planning, probabilistic reasoning, and environment interaction within language models, LATS extends the dynamic capabilities of ReAct~(Reasoning + Acting)~\cite{yao2022react} paradigms, enabling adaptive agent behavior in multi-step problem-solving scenarios. While these approaches show complementary advantages in addressing combinatorial optimization and long-range dependency challenges, computational efficiency and path-pruning strategies remain critical areas for improvement.

\subsection{Progressive Hint Prompting Reasoning}
In the realm of progressive prompting for complex reasoning, the LLMs solve a problem through multiple rounds of messages. Least-to-Most~\cite{least2most} first breaks a problem into several sub-problems and then solve them sequentially. Progressive-Hint Prompting~(PHP)~\cite{zheng2023PHP} advances dynamic problem-solving by fostering iterative, multi-turn interactions between users and LLMs. This method leverages feedback-driven prompts informed by historical outputs to systematically refine reasoning accuracy and coherence. Parallel to this, Cumulative Reasoning~(CR)~\cite{zhang2023CR} emulates human-like incremental cognition by decomposing tasks into structured subtasks and aggregating intermediate results through stepwise integration. Both PHP and CR synergize with foundational frameworks like CoT and its derivatives, collectively strengthening the generation and validation of adaptive reasoning pathways. 

Recent advancements further explore hybrid architectures that combine PHP with retrieval-augmented mechanisms and task-specific distillation. These frameworks aim to balance computational efficiency with robust reasoning fidelity, addressing challenges such as error propagation and context scalability. By integrating iterative feedback loops with external knowledge retrieval, such approaches optimize performance in multi-step reasoning tasks while maintaining generalizability.

\begin{algorithm}[t]
  \caption{RFF-T}\label{rff-with-back}
  \begin{algorithmic}[1]
    \Require  LM $p_{\theta}$, input $x$, max Steps $L$, last step generator $G()$, stepwise reasoner $R()$, state checker $C()$, current state $\{S\}$, target state $\{T\}$, avoid attempts $\{A\}$, verifier $V()$, Output function $O()$
    \State $S_0 \gets x, T_0 \gets t, A_0 \gets \{\}, i \gets 0$
    \While{$i <= L$}
        \State $i \gets i+1$
        \State $A_i \gets \{\}$
        \State $T_i \gets G(p_{\theta}, S_{i-1}, T_{i-1})$
        \State $S_i \gets R(p_\theta, S_{i-1}, T_i, A_{i-1})$
        \If{$C(S_i, T_i)==True$}
            \State $j\gets V(S_i, T_i)$
            \If{$j==i$}
                \State \textbf{break}
            \EndIf
            \State $A_j \gets A_j \cup \{S_j,T_j\}$
            \State $i \gets j$
        \EndIf
    \EndWhile
    \State \Return{$O(p_\theta,x,t|S_i)$}
  \end{algorithmic}
\end{algorithm}

\section{Methods}
Reason from Future~(RFF) is a reasoning paradigm that allows models to solve a question by using forward and backward reasoning alternately. 
We use $p_\theta$ to denote a LLM with parameters $p_\theta$, and $x,t$ to denote the $input$ and $question$. The $\{S\} \sim \{S_0, S_1...S_i\},\{T\} \sim \{T_0, T_1...T_i\}$ denote the list of current states and the list of target states in each step $i$. We define $O(p_\theta,x,t|S_i)$ as the output of the LLM with parameters $p_\theta$ using a prompt consisting of input $x$, target $t$, and hints $S_i$.
In the $i-th$ step, the model identifies the preceding step closest to the current target state $T_{i-1}$ and considers it as the new target state $T_i$ and provides the calculation relationship between the two. Then the model takes the $T_{i-1}$ as the new target for one-step forward reasoning. The model then repeats this step until the latest target state has been achieved~$(S_i = T_i)$. A specific RFF pipeline should consist of three components: 1: Last Step Generator~$G()$; 2: Stepwise Forward Reason~$R()$; 3: State Check~$C()$.

\subsection{Last Step Generator} RFF implements backward reasoning by generating the last previous step. To be specific, RFF decomposes one target state~$T_i$ with current state $S_i$ into a pre-target state $T_{i+1}=G(p_{\theta}, S_i,T_i)$ at a time, the form of the specific sub-target state depends on the on the target of the task, such as a set of numbers (Game of 24), the variables to be found (mathematical problems). It is worth noticing that the transition step between pre-target state~$T_{i+1}$ to target~$T_i$ should be output explicitly to guarantee the correctness of the target decomposition to a certain extent.

\subsection{Stepwise Forward Reason} We consider two different strategies: RFF-T in Algorithm \ref{rff-with-back} and RFF-G in Algorithm \ref{rff-without-back}, to generate the next forward reasoning step for different types of target:

(a) RFF-T: For problems like Game of 24 or Maze game, whose solution is one branch of a searching tree, the model should avoid repeating the wrong attempts in the same layer of the searching tree. We use $\{A\} \sim \{A_0, A_1... A_i\}$ to denote the attempts should be avoid in step $i$, thus the next state should be $S_i \gets R(p_\theta, S_{i-1},T_i, A_{i-1})$.

(b) RFF-G: For the problem like mathematical problems, whose solution is a directed acyclic graph, all the information calculated by the previous states are either useful or redundant but not harmful, so the reasoning path should consider all the information calculated by the previous states, which is 
$S_i \gets S_{i-1} \cup R(p_\theta, x, S_{i-1}, T_i)$.

\begin{algorithm}[h]
  \caption{RFF-G}\label{rff-without-back}
  \begin{algorithmic}[1]
    \Require  LM $p_{\theta}$, input $x$, max Steps $L$, last step generator $G()$, stepwise reasoner $R()$, state checker $V()$, current state $\{S\}$, target state $\{T\}$, Output function $O()$
    \State $S_0 \gets x, T_0 \gets t_0$
    \For{$i = 1$ to $L$}
        \State $T_i \gets G(p_{\theta}, S_{i-1}, T_{i-1})$
        \State $S_i \gets S_{i-1} \cup R(p_\theta, S_{i-1}, T_i)$
        \If{$V(S_i, T_i)==True$}
            \State \textbf{break}
        \EndIf
    \EndFor
    \State \Return{$O(p_\theta,x,t|S_i)$}
  \end{algorithmic}
\end{algorithm}

\subsection{State Check} 

State Check~$C()$ maintains an inference boundary that determines the termination conditions of the inference paradigm. Similar to Stepwise Forward Reason, we set two different strategies to check whether the reasoning comes to the boundary:

(a) RFF-T: For the reason only the correct reasoning path will be saved in the end, the $C(p_\theta, S_i, T_i)$ only considers whether the current state~$S_i$ coincides with the latest target state~$T_i$, or whether the current state requires only one mathematical or logical operation to reach the target state(e.g. present state:(2 3 4), target state:(4 6) in Game of 24). Meanwhile, because RFF-T need to revisit the previous state to explore the thought space, a Verifier $V(S_i, T_i)$ is set to verify whether this path is the correct path when the reasoning comes to the boundary. If the path is a wrong path $V()$ will return the previous state $j$ which should be revisited and record the wrong attempt $(S_j, T_j)$.

(b) RFF-G: Different from the RFF-T, each step of reasoning generates a useful node of the directed acyclic graph, so $C(p_\theta, S_i, T_i)$ considers whether the information the target state needs has already been solved or is noted in the background.

\section{Experiment}

We evaluate the effectiveness of RFF on some widely used LLM reasoning benchmarks, like GAME of 24 and GSM8k. Considering that successful paradigm may be due to the strength of the model itself rather than the strength of the paradigm, leading to difficulty in migrating them to weak models or small models, we carry out our experiments using Llama3-8B-Instruct~\cite{dubey2024llama} and Qwen2.5-7B-Instruct~\cite{yang2024qwen2} as the base models, and more detailed parameters will be shown in specific tasks setup.

\subsection{Game of 24}
The task of Game of 24 originates from ~\citep{yao2024tree}, where the goal is to use four numbers with basic arithmetic operations (+-*/) to obtain 24, and each number can be used only once. 

\begin{figure}[h]
    \centering
    \hspace*{\fill}
    \includegraphics[width=0.5\textwidth]{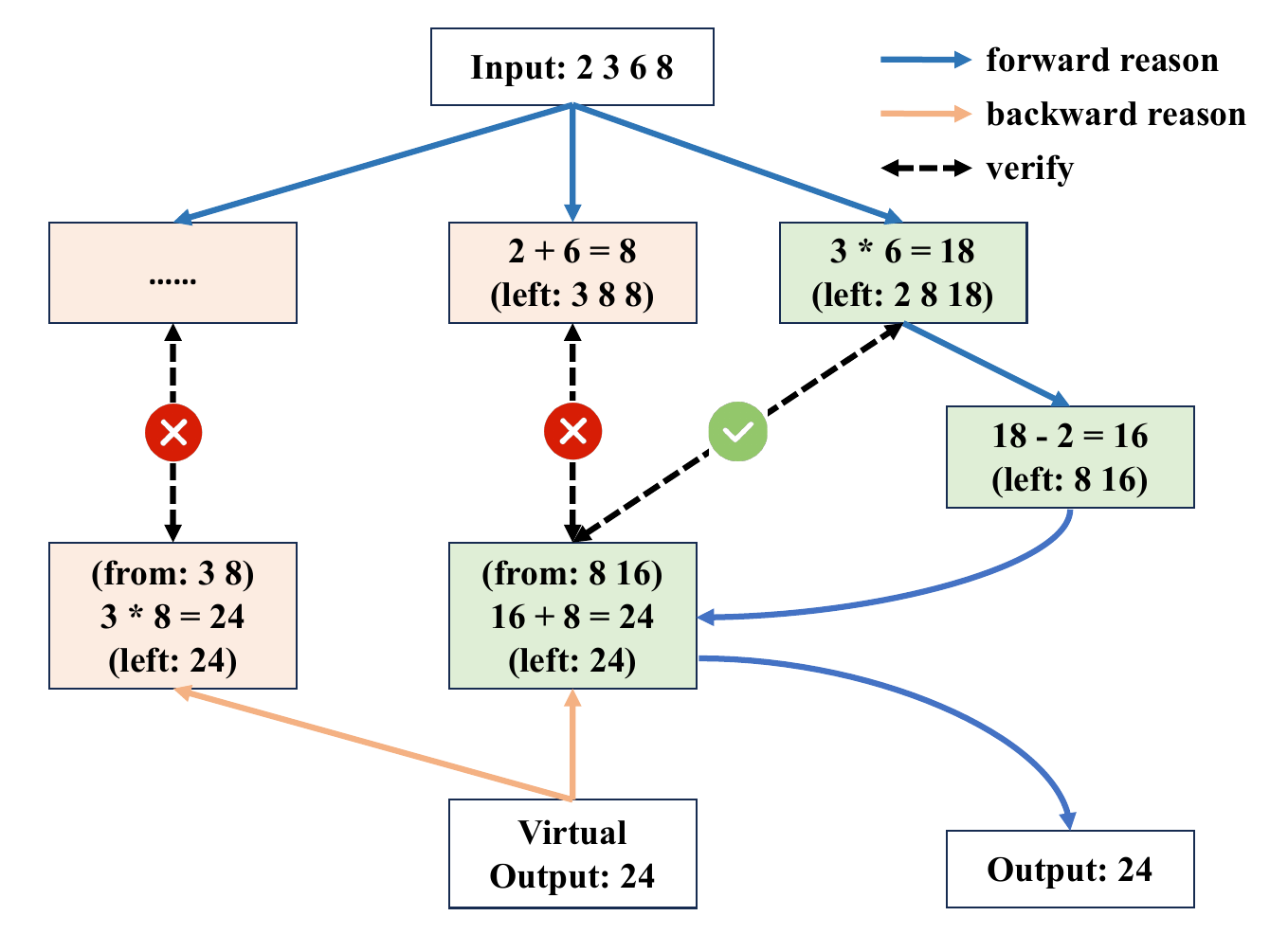}
    \hspace*{\fill}
    \caption{An example of how RFF-T works in Game of 24. Different reasoning paths represent the backtracking mechanism of RFF-T}
    \label{fig:24-example}
\end{figure}

\subsubsection*{Task Setup}
We conduct Game of 24 on Llama3-8B-Instruct with a temperature of 0.7~(consistent with the setup of CoT~\cite{wei2022chain} and ToT~\cite{yao2024tree}). We apply RFF-T because Game of 24 is usually viewed as fetching a branch of the searching tree and is consistent with the paradigm of RFF-T. A solving example can be seen in Figure \ref{fig:24-example}. We conduct 100 times about the middle hard 100 puzzles from 901 to 1000~\cite{yao2024tree}. We also consider each branch of the searching tree as a visit state and record the average visit states of different paradigms, which is proportional to the search space and the consumption of computation.

\begin{table}[H]
    \setlength{\tabcolsep}{4pt}
  \centering
  \begin{tabular}{llll}
    \toprule
    \textbf{Model} & \textbf{Method}           & \textbf{ACC} & \textbf{Visited States} \\
    \midrule
    \multirow{7}{*}{GPT-4} &CoT         &  3\%                  & 1.0 \\
     &ToT(n=1)    & 45\%                  & -    \\
     &ToT(n=5)    & 74\%                  & 61.2\\
     &AoT         & 71\%$^{*}$             & -\\
     &CR(n=1)     & 84\%                  & 11.7\\
     &CR(n=5)     & 94\%                  & 13.7\\
     &\textbf{RFF}(n=5)    & 95\%  &9.3 \\
    \midrule
    \multirow{2}{*}{Llama3-8B}&CR(n=1)           &  9\%                  & 30.9\\
     &CR(n=5)           & 19\%                  & 89.8\\
    \midrule
    \multirow{2}{*}{Llama3-8B}&\textbf{RFF}(n=5)    & 89\%            & \textbf{9.9}\\
     &\textbf{RFF}(n=10)   & \textbf{96\%}   & 15.0\\
    \bottomrule
  \end{tabular}
  \caption{\label{24results}
    The results of the Game of 24, where $n$ denotes the width of the searching tree. $^{*}$ denotes the result is from \cite{sel2023algorithm}.
  }
\end{table}

\begin{figure*}[t]
    \centering
    \makebox[\textwidth][c]{\includegraphics[width=1.1\linewidth]{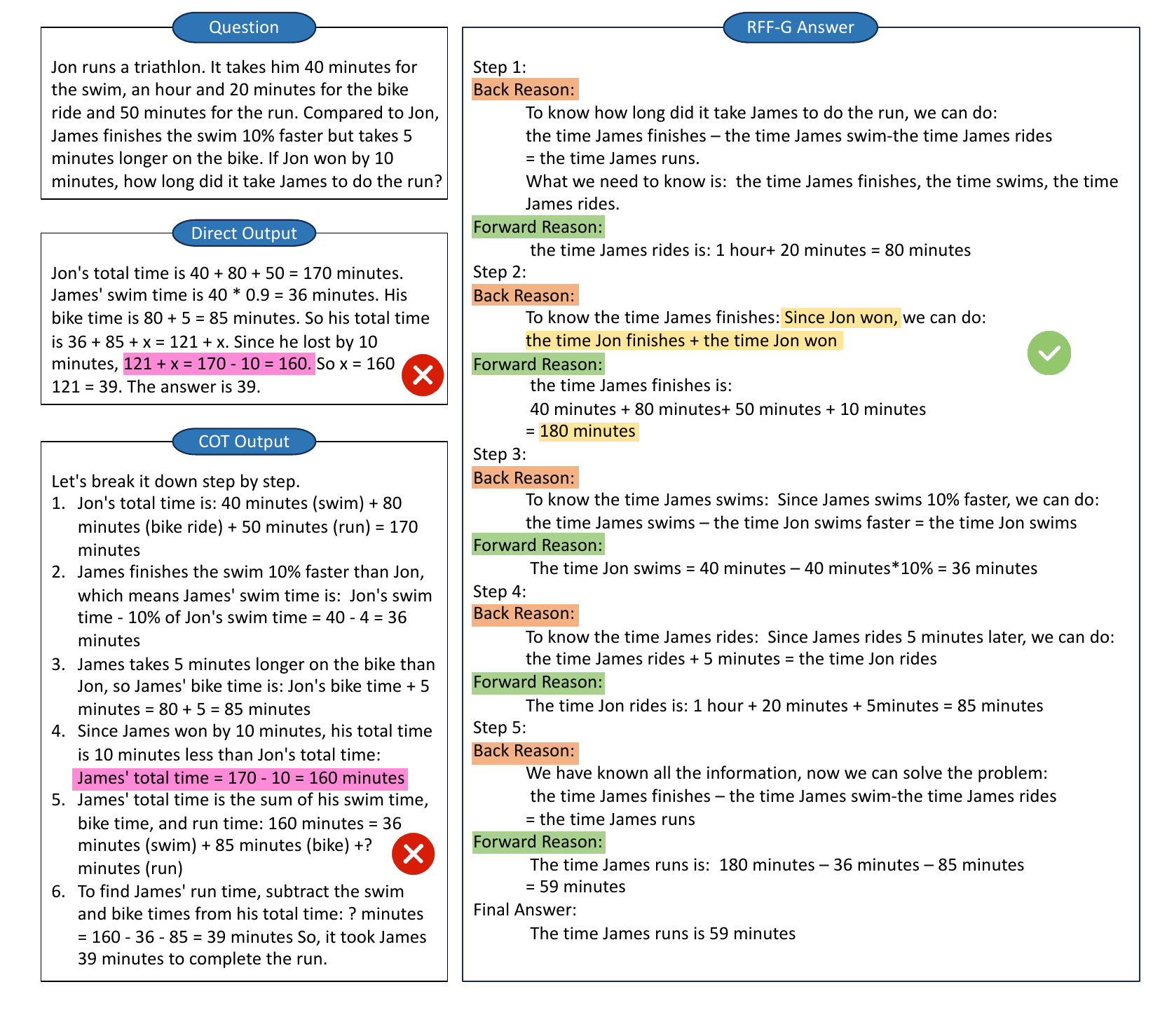}}
    \caption{An example from the GSM8K dataset, with solution generated by Direct, CoT, and RFF paradigms. The former two paradigms tend to connect with "win" to positive operation "more", while RFF will first analyze the background of the "win" and then generate the operation.}
    \label{fig:math-example}
\end{figure*}

\subsubsection*{Baselines}
We employ CoT, ToT, AoT and cumulative reasoning~(CR) with different parameters as the baselines. The setup of CoT is consistent with ~\cite{wei2022chain} and ~\cite{yang2024qwen2}, who employ the intermediate calculation process as the reasoning step. As for ToT and CR, we adapt the settings and prompts from ~\cite{zhang2023CR}. All methods are tested 100 times to get the average result, and unless otherwise specified, the temperature of the model is set to 0.7. We also use GPT-4 as the baseline model. Due to the different space exploring paradigms, we treat those with a similar number of search spaces as a same class for comparison instead of similar branches of searching trees.

\subsubsection*{Results}
As shown in Table~\ref{24results}, RFF with Llama3-8B exhibits outstanding performance even compared to GPT-4. CoT performs badly on this task for the searching-tree-like tasks need the model to explore wide solution space. Searching paradigms like ToT and CR achieve better scores than CoT, meanwhile, the ToT method visits more states because of blind searching. RFF reaches the highest accuracy and least visit-states at the same level: when the visit-state is around 10, RFF reaches the best accuracy of 89\% compared to CR with GPT-4 at 84\%; when the visit-state is around 14, RFF reaches an accuracy of 96\% compared to CR with GPT-4 at 94\%. The fewer visit-states and higher accuracy are due to the searching space in RFF being much smaller and reasonable than simply forward searching~(e.g. for "1 2 12 12", with the target "12+12=24", LLM will less likely explore ways like "2+12=14"). 

\begin{table*}[!ht]
  \setlength{\tabcolsep}{9pt}
  \centering
  \begin{tabular}{lllllll}
    \toprule
    \textbf{Model}    &  \textbf{Method} & \textbf{GSM8K} & \textbf{SVAMP}  & \textbf{ASDiv}  & \textbf{MATH} & \textbf{AVG}\\
    \midrule
    \multirow{5}{*}{Llama3-8B-Instruct} 
    & CoT         
    & 75.6\% & 80.5\% & 82.3\% & 32.8\% & 67.8\%\\
    & Least-to-Most 
    & 79.5\% & 86.8\% & 84.4\% & 38.8\% & 72.4\%\\
    & Give-me-Hint
    & 77.3\% & 87.9\% & 86.0\% & 37.0\% & 72.1\%\\
    & CR     
    & 77.0\% & 71.2\% & 84.8\% & 40.2\% & 68.3\%\\
    & \textbf{RFF}
    & \textbf{83.8}\% & \textbf{89.7}\% & \textbf{86.7}\% 
    & \textbf{41.4}\% &\textbf{75.4}\%\\
    \midrule
    \multirow{3}{*}{Qwen2.5-7B-Instruct} 
    & CoT 
    & 87.2\% & 92.1\% & 88.0\%  & 74.6\% & 85.5\% \\
    & CR  
    & 87.7\% & 83.7\% & 91.9\% & 78.2\% & 85.4\% \\
    & \textbf{RFF}         
    & \textbf{89.5}\% & \textbf{95.1}\% & \textbf{92.2}\% 
    & \textbf{79.8\%}  &\textbf{89.1}\%\\
    \bottomrule
  \end{tabular}
  \caption{\label{GSM8Kresults}
    The results of the math problems. The score represents the accuracy of the benchmarks.  The best results in each column is highlighted in \textbf{bold}. The AVG  represents the average value of the four benchmarks.
  }
\end{table*}

\subsection{Math Problem Benchmark}
This task contains four datasets: GSM8K, SVAMP, AsDiv  and MATH-500. GSM8K is a mathematical dataset with 1319 test data, each question requires 3-10 steps of reasoning and calculation. SVAMP and ASDiv are two simple math problem datasets with 1000 and 2096 data respectively, each question requires 1-2 steps of reasoning and calculation. MATH-500 is a subset of 500 mathematical problem from the MATH ~\cite{math} benchmark, which is a much more harder benchmark than GSM8K.

\subsubsection*{Task Setup}
We conduct this task on Llama3-8B-Instruct and Qwen2.5-7B-Instruct with a greedy search to exclude the influence of random numbers on textual reasoning. We apply RFF-G for the mathematical puzzle as its solution can be seen as a directed acyclic graph from the question to the answer. We employ 1 shot as the example to lead the model to perform formatted reasoning.

\subsubsection*{Baselines}
Considering the nature of the math problems, CoT and CR are chosen as baselines for their excellent ability for complex thinking and multi-jump reasoning either. CoT and CR are set with one shot to balance the influence of the same setup in RFF-G. 
CoT generates a continuous chain of thoughts until the model answers the question while CR generates a few hints first, then generates simple questions and answers until the model thinks it's enough to answer the question. 
We also conducted two extra baseline not showing in the Game of 24: Least-to-Most ~\cite{least2most}, which generates the sub-problem pipeline first and then follow the pipeline to solve the problem and Give-me-Hint ~\cite{givemehint}, which generates helpful hints to help LLM solve the problem. Both baselines are recent and typical to serve as convincing control groups.

\subsubsection*{Results}

Table~\ref{GSM8Kresults} shows the accuracy of Llama3-8B-instruct and Qwen-2.5-7B-Instruct on four datasets and our method RFF shows great advantages over the other methods. 
Meanwhile, there is another an interesting phenomenon: since the methods present great improvement of accuracy to CoT on GSM8K, ASDiv and MATH datasets, contributing to the better focus on details and relations about progressive prompting methods. However, CR fails to reach the average level of other methods on simple task SVAMP with 71.2\% compared to 85.1\%, and demonstrates better performance on hard task MATH with 40.2\% to 36.2\%. We carefully check the output of the CR and our RFF to and find that the detailed hints and question-answer pairs can be helpful in hard problems like MATH, but they lead to a harmful phenomenon of overthinking when facing simple problems like SVAMP. And our method RFF, benefiting from the State Checker, can avoid overthinking when the State Checker thinks the reasoning should be stopped. We also notice the gap between RFF and CoT is increasing with the base ability of the model decreasing~(from Qwen to Llama), demonstrating a significant complementary effect on the model's reasoning ability.

\subsection{Commonsense Problem Benchmark}

To further explore the effectiveness of RFF in different NLP tasks, we have conducted experiments on commonsense problem benchmarks. Commonsense problems usually refer to those that require basic knowledge and reasoning ability in human daily life to solve. These problems rely on background knowledge, which is usually not explicitly stated in the problems.
we have conducted experiments on two widely used commonsense benchmarks: CommonQA~\cite{commonqa} and LogiQA~\cite{logiqa} using RFF-G.
Both two benchmarks are multiple-choice, which contains 12102 and 8678 questions respectively, each with one correct answer and four choices.

\subsubsection*{Task Setup}
We conduct this task on Llama3-8B-Instruct to test on this benchmark. Considering the nature of the commonsense tasks, we choose RFF-G to solve the question. We judge the accuracy of the answer by observing whether the model outputs the correct option, any other forms of answers will be viewed as wrong.

\subsubsection*{Baselines}
For commonsense tasks, it will be much helpful if the model is given background informations before answering the question, so the hints based and progressive prompt reasoning paradigms will serve as effective references.
Same as math benchmark, we choose COT, CR, Least-to-Most and CR as our experiment baselines. 

\subsubsection*{Results}
As shown in Table \ref{nlp reasoning}, all the reasoning paradigms achieve a better accuracy on CommonQA, demonstrating the complex prompting can easily improve the performance on it. However, the differences among these methods are not so apparent~(all distributed at around 76\%), so the sole result of CommonQA is not sufficient to express the superiority of our method. In LogiQA, the differences shows a greater gap among these baselines: the results of Least-to-Most and Give-me-Hint are close to COT while the results of RFF and CR shows a significant improvement over COT.

\begin{table}[h]
  \setlength{\tabcolsep}{4pt}
  \centering
  \begin{tabular}{llll}
    \toprule
    \textbf{Method} & \textbf{CommonQA} & \textbf{LogiQA} & \textbf{AVG} \\
    \midrule
    COT & 73.1\% & 41.8\% & 57.5\% \\
    CR & 75.4\% & \textbf{45.5\%} & \underline{60.5\%} \\
    Least-to-Most & 74.3\% & 42.9\% & 58.6\% \\
    Give-me-Hint & \underline{76.6\%}   &   42.0\%  & 59.3\% \\
    \textbf{RFF} &   \textbf{77.1\%}     &   \underline{45.2\%} & \textbf{61.2\%} \\
    \bottomrule
  \end{tabular}
  \caption{\label{nlp reasoning}
    The results of commonsense reasoning benchmarks. The best result in each column is highlighted in \textbf{bold}, and the second best in each column is added an underline.
  }
\end{table}

The results also indicate that in commonsen tasks, especially in easy task~(CommonQA), simple prompting method~(Give-me-Hint) achieves better scores than complex prompting methods~(Least-to-Most and CR), which we think it’s for the over thinking of these methods. Meanwhile, RFF can still maintain a high score due to its step-evaluation mechanism: when facing simple questions, the forward reasoning can quickly meets the backward reasoning, then the RFF turns to simple COT to solve the problems instead of over thinking.

\subsection{Studies of Redundant Thinking}
We investigate the limitations of traditional algorithms for solving Game of 24. While conventional breadth-first searching methods perform well in low-dimensional solution spaces, their unguided exploration mechanisms may lead to significant computational resource waste and efficiency degradation when handling higher-dimensional problems. 

To validate this theoretical hypothesis, we constructed an experimental dataset comprising 100 enhanced problems (IDs 901-1000) by strategically adding the constant "1" to original four-number combinations, creating five-number variants. Theoretically, this operation preserves the solvability of problems (based on arithmetic identity transformations) and is expected to decrease the difficulty by introducing a redundant variant.

\begin{table}[h]
\setlength{\tabcolsep}{4pt}
  \centering
  \begin{tabular}{llll}
    \toprule
    \textbf{Model} & \textbf{Method}           & \textbf{ACC} & \textbf{Visited States} \\
    \midrule
    \multirow{3}{*}{GPT-4} &CR(n=5)   & 76\%  & 7.06\\
    &\textbf{RFF}(n=5)   & 89\%  &\textbf{5.96}\\
    &\textbf{RFF}(n=10)   & \textbf{93}\%  & 9.13\\
    \midrule
    \multirow{3}{*}{Llama3-8B}&CR(n=5)         & 26\%   & 96.56\\
    &\textbf{RFF}(n=5)    & 85\%      & \textbf{28.62}\\
    &\textbf{RFF}(n=10)   & \textbf{92}\%      & 56.13\\
    \bottomrule
  \end{tabular}
  \caption{\label{24results-5}
    The results of 5 numbers of the Game of 24.
  }
\end{table}

\begin{figure*}[t]
    \centering
    \makebox[\textwidth][c]{\includegraphics[width=1.1\linewidth]{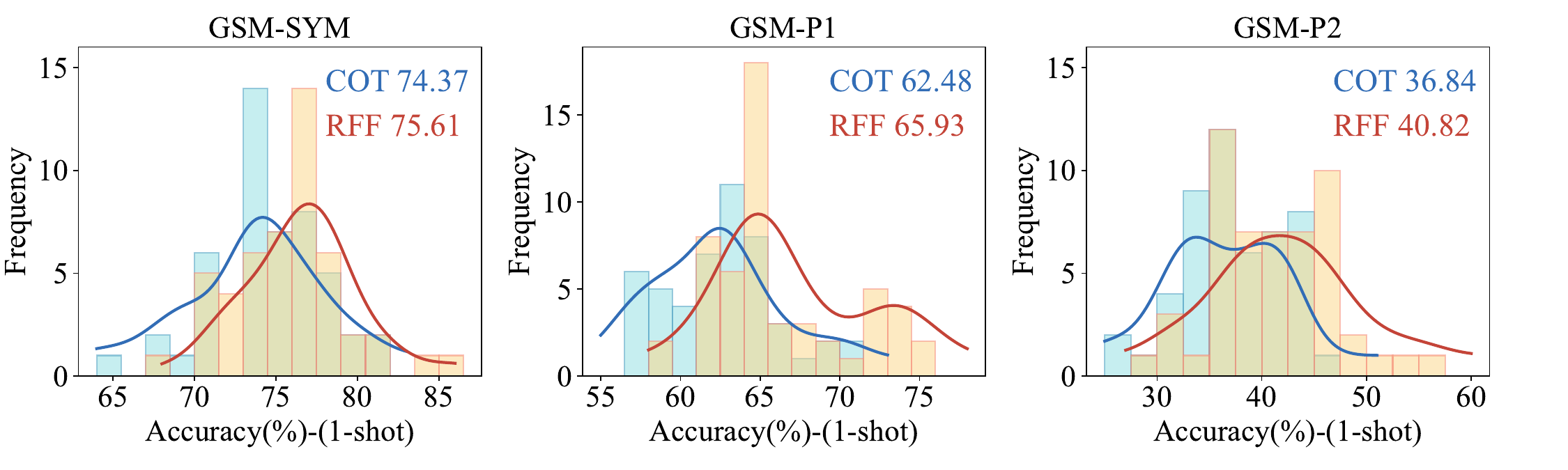}}
    \caption{The result of CoT and RFF on GSM-Symbolic dataset. The score in the upper right of the chart is the average accuracy score of the 50 subsets. RFF shows more stable and better distributed at high accuracy.}
    \label{fig:sym_result}
\end{figure*}

As shown in Table~\ref{24results-5}, after adding a redundant "1", the performance of CR decreased significantly with GPT-4~(from 94\% to 76\%). In contrast, RFF achieves a higher success rate with fewer visit states compared with CR. When we further expand the search space, the model's performance continues to improve. At the same time, we observe the smaller model requires more attempts to achieve performance comparable to the original data. However, RFF consistently surpasses CR in terms of success rate and resource consumption. The result demonstrates the effective prospective state space pruning in RFF and validates the superiority of the searching space convergence mechanisms of RFF.

\subsection{Studies of Robust Thinking}
To address the data leakage risks associated with the widespread adoption of the GSM8K, researchers~\cite{mirzadeh2024gsm} proposed the GSM-Symbolic benchmark through semantic-preserving transformations. This dataset generates derivative problems via entity/quantity substitution~(GSM-SYM) and the addition of single~(GSM-P1) or dual~(GSM-P2) conditional constraints to the original question. Despite theoretical expectations that surface-level modifications~(e.g., name/quantity changes) should not impact reasoning capabilities, empirical observations reveal significant accuracy degradation across all models. Following the standard evaluation protocol of GSM8K, we systematically assess the reasoning generalization of RFF models using the publicly available SYM, P1, and P2 subsets~(each subset contains 50 variants from an original dataset). We employ Llama3-8B-Instruct with CoT and RFF methods to conduct this task.

Figure~\ref{fig:sym_result} shows the accuracy distribution of the 50 variants datasets in three subsets. The result exhibits that the accuracy of both methods drops on these three datasets, representing the fragility of the reasoning ability of models. 

However, RFF still has advantages in the average accuracy and shows a more concentrated and more accurate distribution. The result emphasizes that the form of forward reasoning guided by backward reasoning is quite robust in the face of variant problems.

\section{Conclusion}
In this paper, we introduce Reason from Future~(RFF), a novel reasoning paradigm aiming at enhancing the reasoning ability of LLMs for complex problems. RFF leverages a bidirectional reasoning framework integrates top-down planning with bottom-up reasoning accumulation to generate a solution path. This aids in the convergence of the search space for the model, thereby enhancing inference efficiency. Simultaneously, it allows the model to focus on critical information, which improves the accuracy of reasoning. RFF has demonstrated superior performance across both searching tree tasks~(Game of 24) and directed acyclic graph tasks~(math problems and commonsense problems), showing the potential to enhance the model's reasoning capabilities.

\section*{Limitations}
The effectiveness of RFF relies on the model's ability for reverse thinking. Since the model has not been trained with specialized data, there can be rare instances where errors in the final step of reverse reasoning lead to failure. In future work, we will introduce fine-tuning or reinforcement learning to further enhance the generalizability of this reasoning paradigm.

\section*{Acknowledgments}
This research was partially supported by National Natural Science Foundation of China under Grant No.12326612, No.82202984, Zhejiang Key R\&D Program of China under Grant No.2024SSYS0026, Zhejiang Key Laboratory of Medical Imaging Artificial Intelligence, and the Transvascular lmplantation Devices Research Institute (TIDRI).

\bibliography{custom}

\appendix

\begin{figure*}[t]
    \centering
    \includegraphics[width=1.0\textwidth]{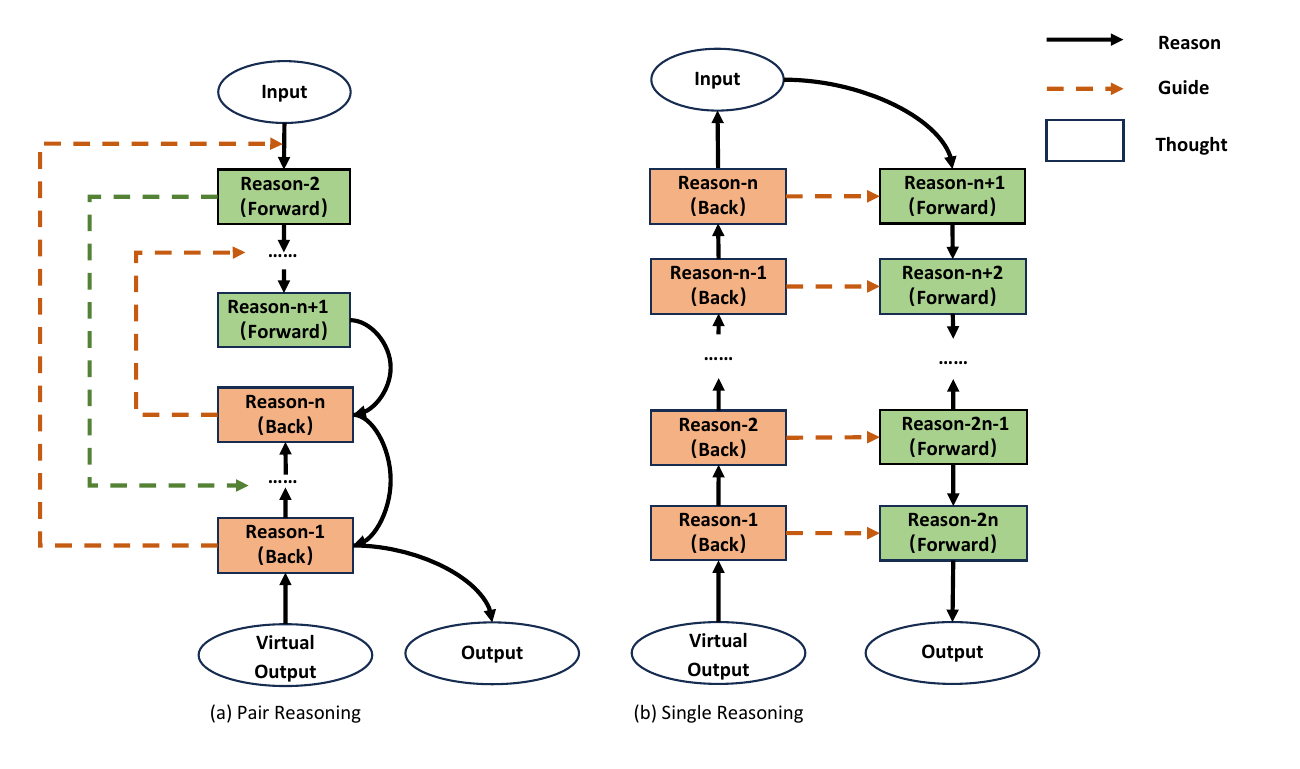}
    \caption{Two different strategies of backward reasoning}
    \label{fig:enter-label}
\end{figure*}

\section{Detailed Difference between Baselines}
Although most progressive prompting methods employ the reasoning paradigms by dividing a hard problem into several simple subproblems, the mechanism for decomposing the problem and the mechanism for how to solve the subproblems are different and serves as the core of the reasoning paradigms.
We carefully compare the two typical baselines we used in our work~(Least-to-Most and CR), and analyzed their characteristics and difference with RFF.

\subsection{Difference between Least-to-Most}

For Least-to-Most, it generates its sub-problems of the final problem in the very beginning, then the LLM follows the pipeline of the sub-problems to generate its reasoning. It’s a two-stage plan-solve reasoning and the sub-problems never change during the solution, so the sub-problems may be wrong or insufficient because of a lack of the intermediate information generated by the reasoning process.

The biggest difference between RFF and Least-to-Most is that process of RFF starts from the end state of the problem, RFF doesn’t plan just at the beginning but to continue to generate questions or hints or guidance by backward reasoning to lead LLM to better reason forward. 

\subsection{Difference between CR}

For CR, it generates several hints for the problem first, the hints are helpful to solve the questions. Then the LLM continues to generate subproblem which based on the hints and then answer it until the LLM can solve the problem.

Differ from CR, which generates guidances from the previous visiting states to reduce hallucinations, our method~(RFF) generates guidance from the backward reasoning to give the model a full-perspective of the problem. Additionally, the detailed hints and nonstop subproblems may lead CR to overthinking, meanwhile, the full-perspective and State Check ensure the ability of RFF to stop overthinking quickly when facing simple questions with complex prompting.

\section{Extra Experiments on Reasoning Paradigms}

We compare two different backward reasoning strategies on Math Problems called Pair Reasoning (same as RFF in \ref{fig:framework}) and Single Reasoning as shown in Figure \ref{fig:enter-label}, we conduct experiments on GSM8K dataset using Llama3-8B-Instruct with greedy search.

\begin{table}[h]
\setlength{\tabcolsep}{7pt}
  \centering
  \begin{tabular}{lll}
    \toprule
    \textbf{Model} & \textbf{Method} & \textbf{ACC}\\
    \midrule
     \multirow{3}{*}{Llama3-8B} & CoT& 75.6\%\\
     & Pair Reasoning RFF                 & \textbf{83.8}\%\\
     & Single Reasoning RFF               & 69.8\%\\
    \bottomrule
  \end{tabular}
  \caption{\label{appendix_result}
    The results of different back reasoning strategies. 
  }
\end{table}

As shown in Table \ref{appendix_result}, the performance of Single Reasoning RFF drops badly on GSM8K dataset and even worse than CoT, we assume that the weakness of the strategy of Single Reasoning is that when deducing the whole chain of backward thought, the situations of multi-hop seriously affects the backward reasoning without new information generated by the forward reasoning. So when employing backward reasoning paradigms, a continuous generation of new information is needed.

\section{Appendix for Prompts}
The design of prompts is critical to lead the model to reason exactly according to the paradigm we have planned. We design these prompts deliberately to ensure the model reason and output according to the format we give it.

\begin{figure*}[h]
    \centering
    \includegraphics[width=1.0\textwidth]{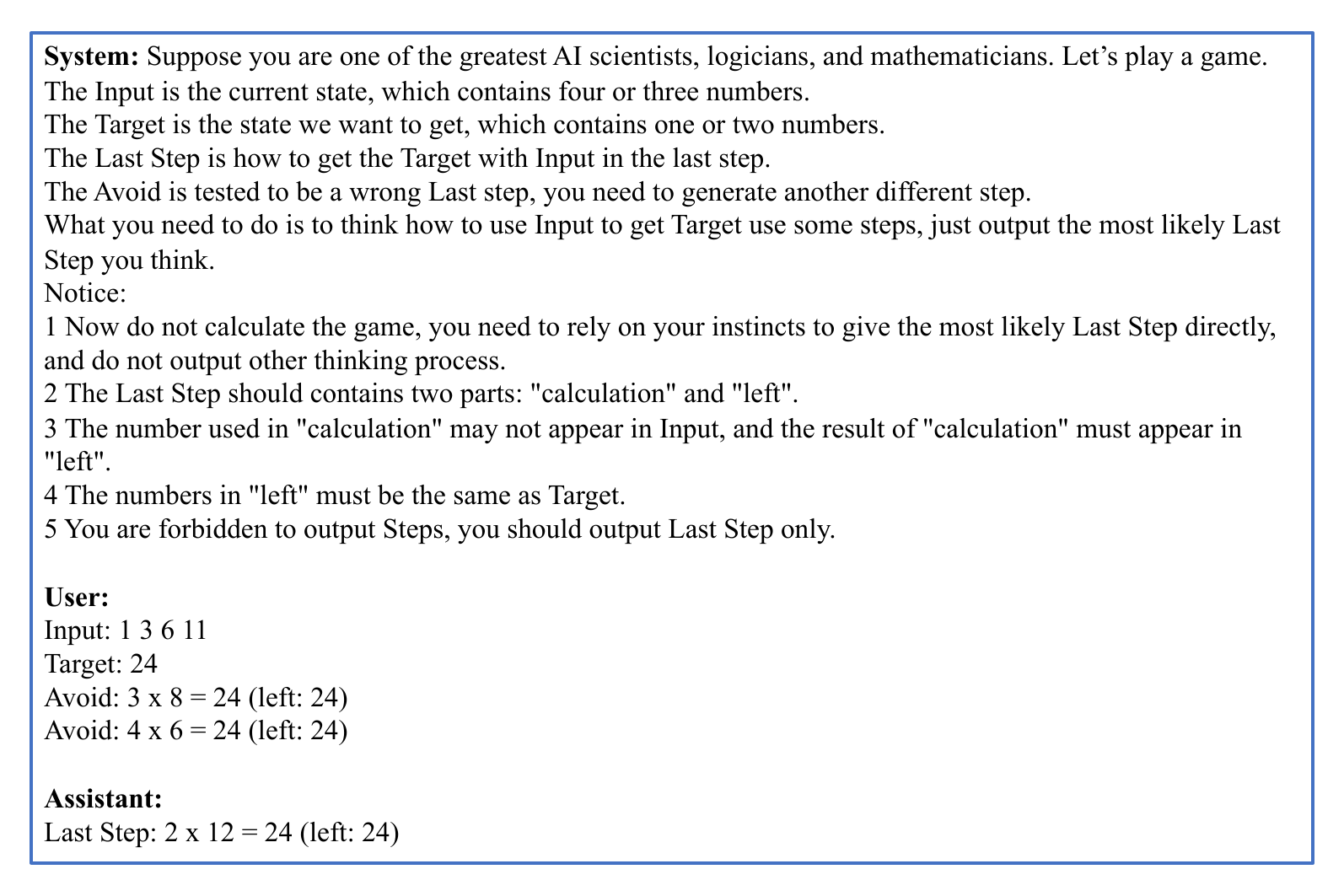}
    \caption{Prompts of Last Step Generator for Game of 24}
    \label{Prompts of Last Step Generator of Game of 24}
\end{figure*}

\begin{figure*}[h]
    \centering
    \includegraphics[width=1.0\textwidth]{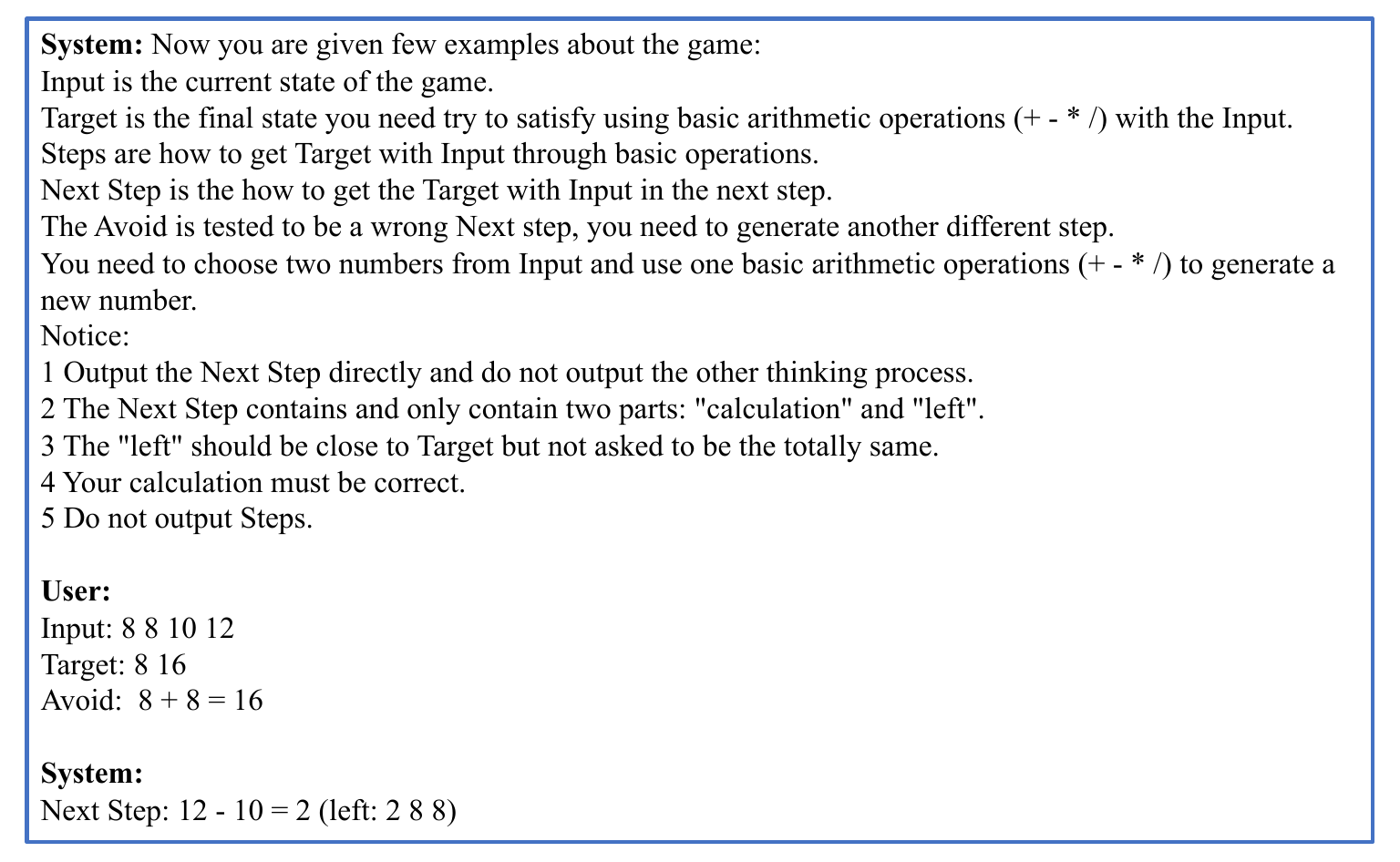}
    \caption{Prompts of Stepwise Forward Reason for Game of 24}
    \label{Prompts of Stepwise Forward Reason of Game of 24}
\end{figure*}

\begin{figure*}[h]
    \centering
    \includegraphics[width=1.0\textwidth]{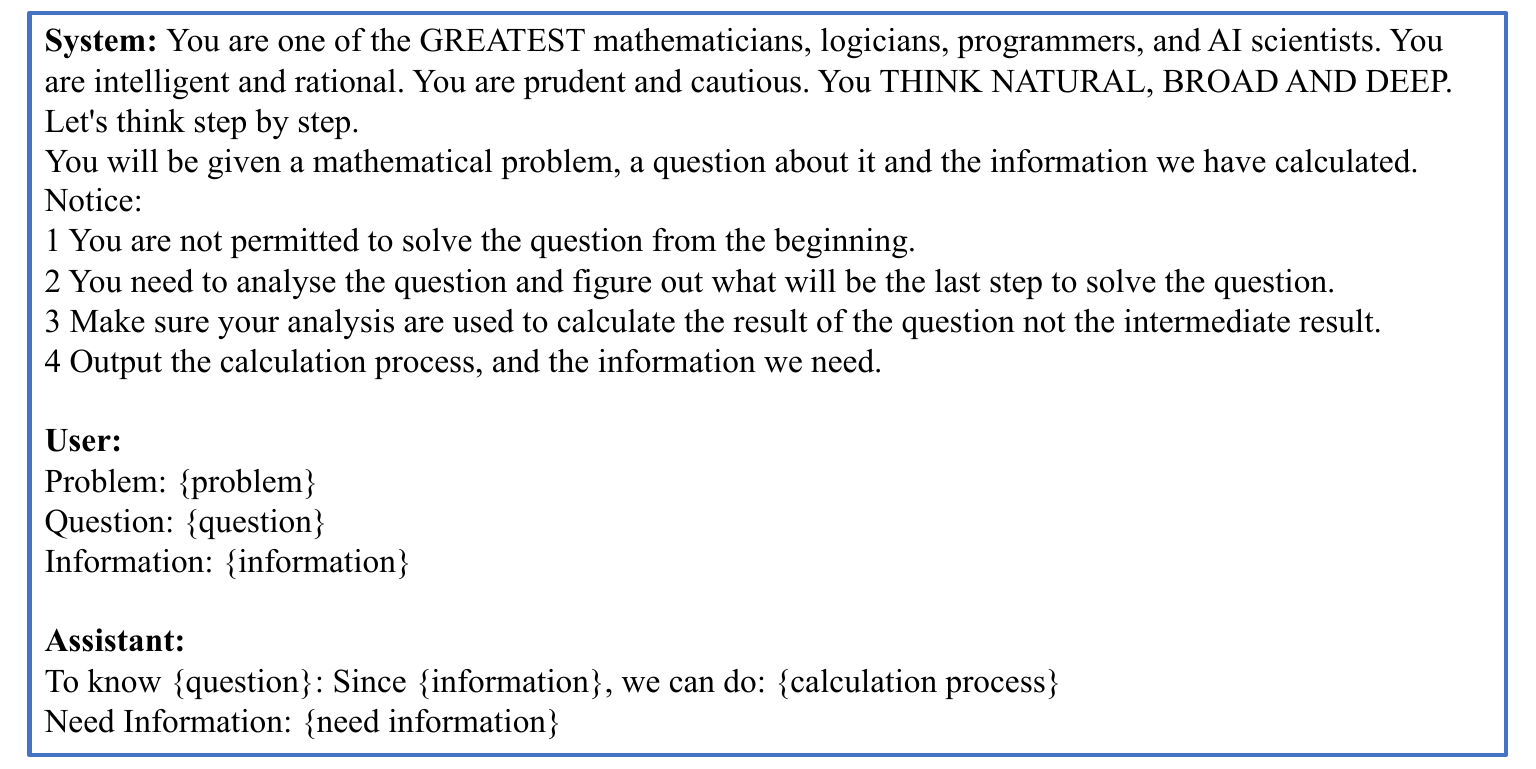}
    \caption{Prompts of Last Step Generator for math problems}
    \label{Prompts of Last Step Generator of math problems}
\end{figure*}

\begin{figure*}[h]
    \centering
    \includegraphics[width=1.0\textwidth]{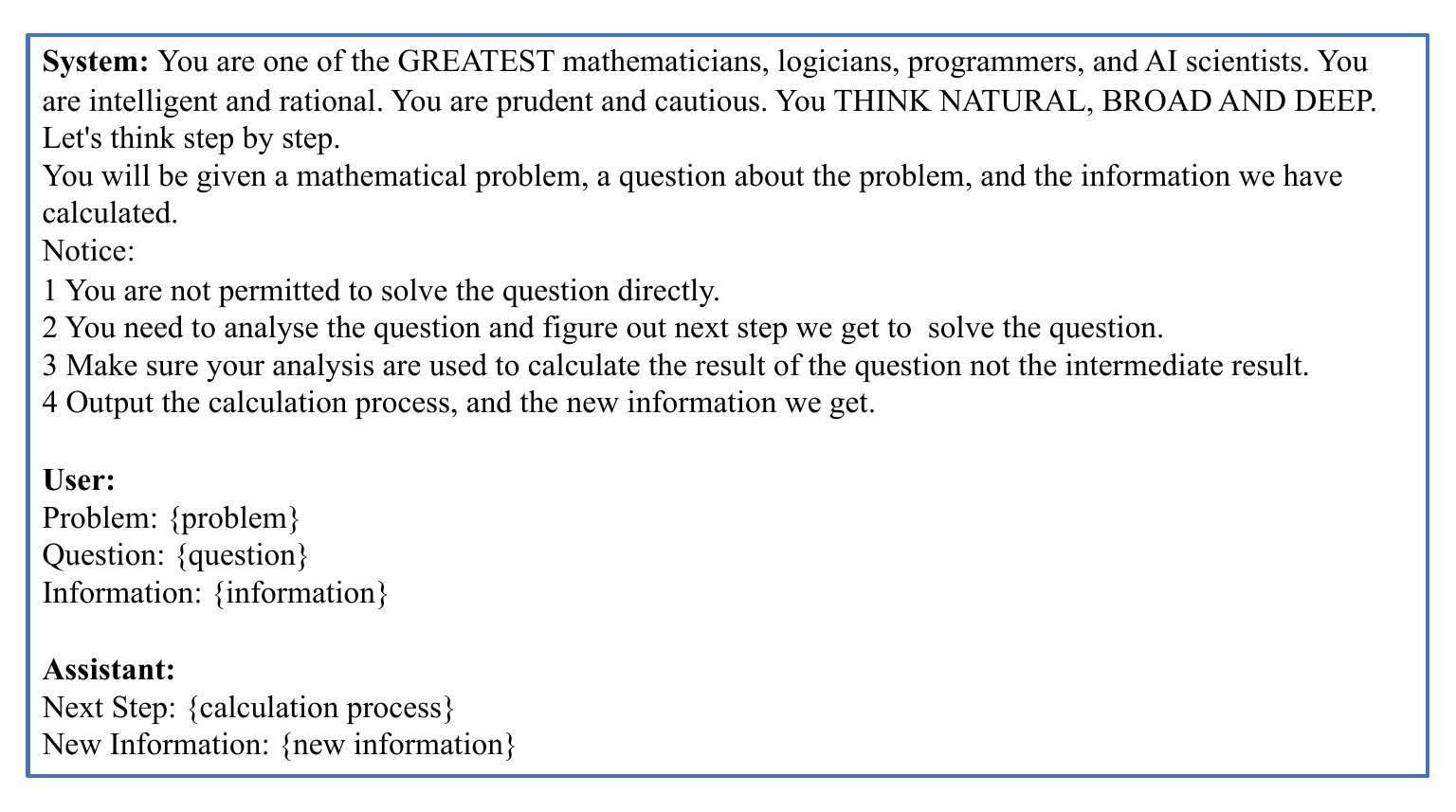}
    \caption{Prompts of Stepwise Forward Reason for math problems}
    \label{Prompts of Stepwise Forward Reason of math problems}
\end{figure*}

\begin{figure*}[h]
    \centering
    \includegraphics[width=1.0\textwidth]{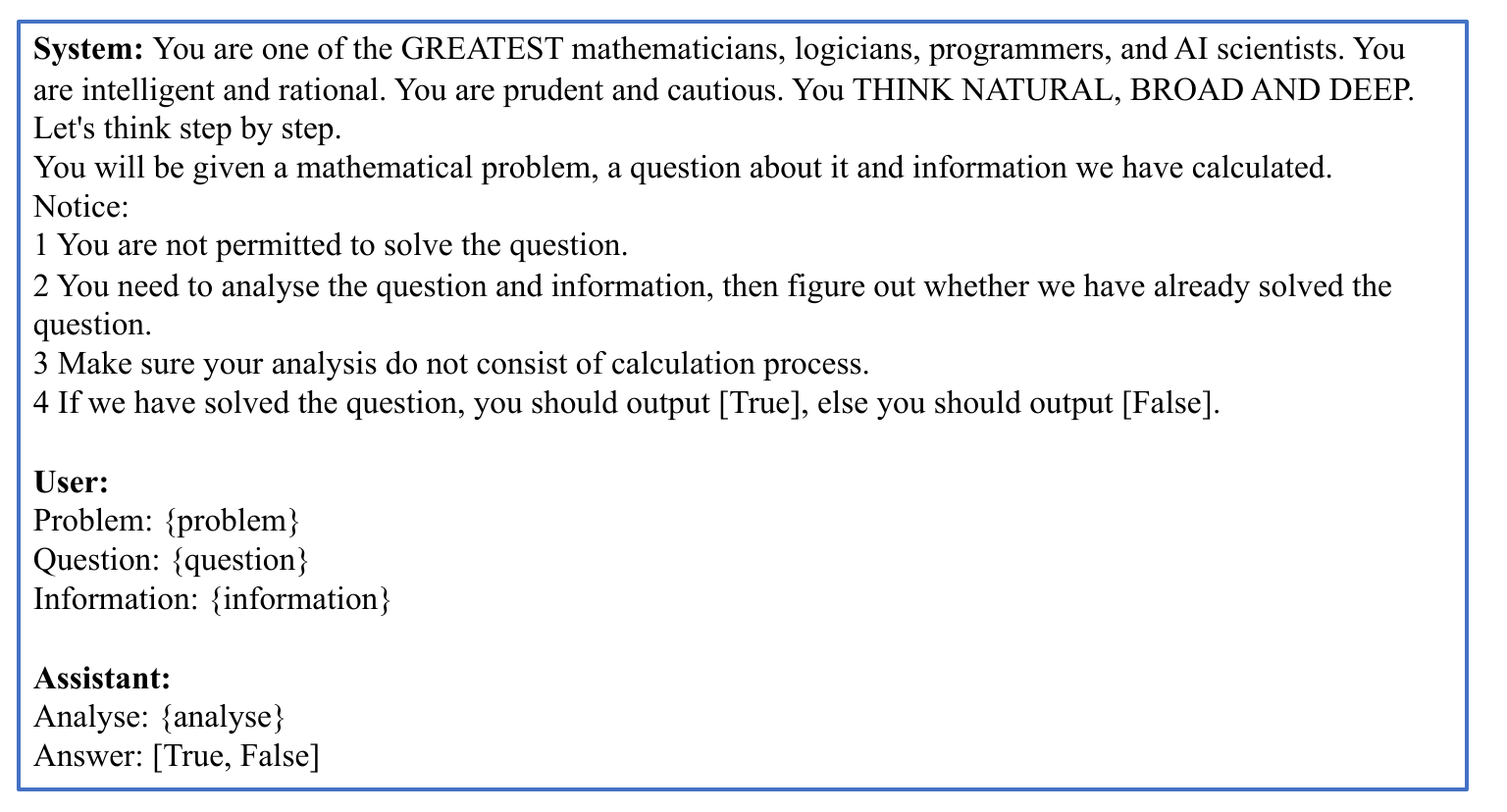}
    \caption{Prompts of State Check for math problems}
    \label{Prompts of State Check of math problems}
\end{figure*}

\end{document}